\documentclass[manuscript, screen]{acmart}

\usepackage{colortbl}
\usepackage{multirow}
\usepackage{graphicx}

\renewcommand\footnotetextcopyrightpermission[1]{} 
\AtBeginDocument{%
  \providecommand\BibTeX{{%
    \normalfont B\kern-0.5em{\scshape i\kern-0.25em b}\kern-0.8em\TeX}}}

\newcommand{\hide}[1]{}

\newcommand{\vect}[1]{{#1}}
\newcommand{\matr}[1]{{#1}}

\newcommand{\method}{\texttt{ROBAD}} 

\setcopyright{none} 

\begin{document}


\title{ROBAD: Robust Adversary-aware Local-Global Attended Bad Actor Detection Sequential Model}

\author{Bing He}
\affiliation{%
  \institution{Georgia Institute of Technology}
  \country{USA}}
\email{bhe46@gatech.edu}

\author{Mustaque Ahamad}
\affiliation{%
  \institution{Georgia Institute of Technology}
  \country{USA}}
\email{mustaque.ahamad@cc.gatech.edu}

\author{Srijan Kumar}
\affiliation{%
  \institution{Georgia Institute of Technology}
  \country{USA}}
\email{srijan@gatech.edu}



\begin{abstract}

Detecting bad actors is critical to ensure the safety and integrity of internet platforms. Several deep learning-based models have been developed to identify such users. 
These models should not only accurately detect bad actors, but also be robust against adversarial attacks that aim to evade detection. 
However, past deep learning-based detection models do not meet the robustness requirement because they are sensitive to even minor changes in the input sequence. 
This allows bad actors to evade detection by carefully crafting a sequence of posts as new inputs, degrading the efficacy of these models. 
To address this issue, we focus on 
(1) improving the model understanding capability such that even when the input sequence varies, the detector can still make the same prediction of the user; 
(2) enhancing the model knowledge such that the model can recognize potential input modifications when making predictions. 
To achieve these goals, we create a novel transformer-based classification model, called ROBAD (RObust adversary-aware local-global attended Bad Actor Detection model), which uses the sequence of user posts to generate user embedding to detect bad actors. 
Particularly, ROBAD first leverages the transformer encoder block to encode each post bidirectionally, thus building a post embedding to capture the local information at the post level. 
Next, it adopts the transformer decoder block to model the sequential pattern in the post embeddings by using the attention mechanism, which generates the sequence embedding to obtain the global information at the sequence level. 
Finally, to enrich the knowledge of the model when it processes embeddings of original sequences, embeddings of modified sequences by mimicked attackers are fed into a contrastive-learning-enhanced classification layer for sequence prediction. 
In essence, by capturing the local and global information (i.e., the post and sequence information) and leveraging the mimicked behaviors of bad actors in training, ROBAD can be robust to adversarial attacks. 
We conduct extensive experiments on two real-world datasets (Yelp and Wikipedia, both with ground-truth of bad actors) and the results show that ROBAD can effectively detect bad actors when under state-of-the-art adversarial attacks. 
In particular, ROBAD outperforms representative compared methods by having the highest F1 score when no attack and the lowest F1 drop when under attack. 
This demonstrates the superiority of ROBAD.
Overall, this work paves the path toward the next generation of robust bad actor detection models.

\end{abstract}

\settopmatter{printfolios=true} 


\maketitle

\section{Introduction}

\begin{figure}[!htbp]
    \centering
    \includegraphics[width=0.8\columnwidth]{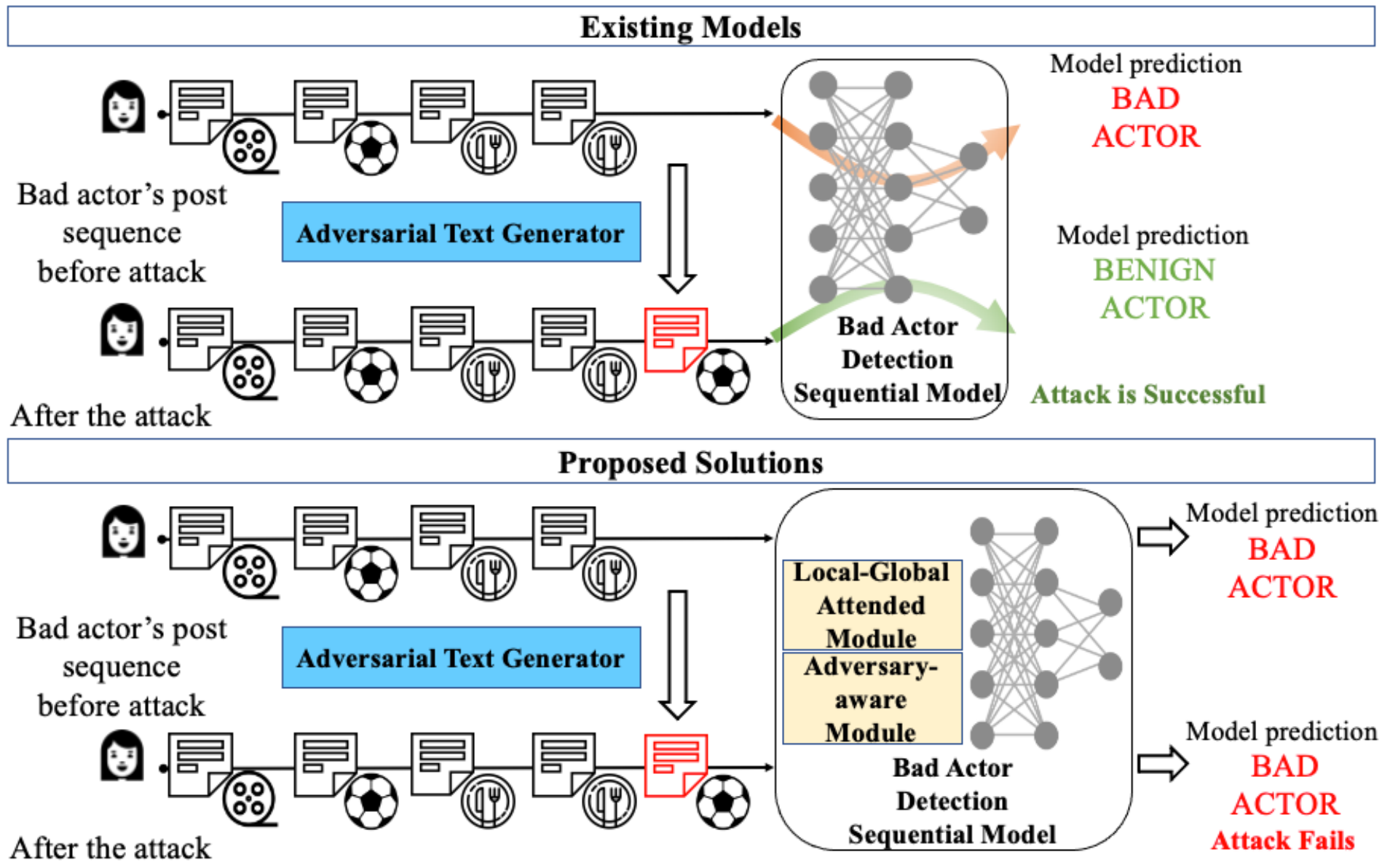}
    \caption{
    Bad actor detection sequential models are always used to identify bad actors. However, the next post attack by an adversary can lead the same model to misclassify it as a benign actor (top row). 
    Our proposed solution, \method{}, built on the local-global attended and adversary-aware modules can robustly identify bad actors even under attack (bottom row).
    }
    \label{fig:setting}
\end{figure}

The popularity of online platforms, e.g., social media and e-commerce sites, has unfortunately made them prime targets for bad actors seeking personal gain~\cite{noorshams2020ties, kumar2018rev2, kumar2017army}. 
This has led to a surge in undesirable activities — including fake user accounts~\cite{noorshams2020ties}, spammers~\cite{dou2020robust, rayana2015collective}, purveyors of fake news~\cite{shu2017fake, le2020malcom}, abnormal users~\cite{chalapathy2019deep}, vandal editors~\cite{kumar2015vews}, fraudsters~\cite{kumar2018rev2}, and sockpuppets~\cite{kumar2017army} — degrading the safety and integrity of online communities. 
For instance, it was reported that approximately 5\% of Facebook's monthly active users in 2019 were identified as fake accounts~\cite{noorshams2020ties}. Similarly, an alarming 63\% of reviews on beauty products on Amazon were found to be posted by fraudulent users~\cite{teodora2021}. 
Given these statistics, identifying such bad actors becomes paramount to maintaining a secure and trustworthy online environment.

Deep learning models have become popular for detecting bad actors, as exemplified by the TIES model employed by Facebook~\cite{noorshams2020ties}. 
These sequential models, termed deep user sequence embedding-based classification models, first deploy neural networks to create user embeddings based on the sequences of their posts. 
These embeddings are then fed into a linear classifier layer to predict user behavior. 
For instance, as illustrated in Figure~\ref{fig:setting}, a model trained on the sequence of a user's posts can be adept at identifying bad actors. 

However, existing research has shown that these deep learning solutions can be vulnerable to adversarial attacks~\cite{he2021petgen}. As depicted in Figure~\ref{fig:setting}, the bad actor can draft a new post, and the entire user sequence is misclassified as benign by the same classifier. 
More alarmingly, ~\citet{he2021petgen} found that even under the simplest attack where a historical post is copied and appended to the original sequence, the F1 score drops by 12.67\%, let alone the sharpest drop of 21.41\% through the optimized attack~\cite{he2021petgen}. 
Thus, improving model robustness against such manipulation is crucial.

Building a robust classification model is non-trivial because deep sequential models are sensitive to even minor changes in the input.  It finally leads to the vulnerability of the model~\cite{he2021petgen}.  
More critically, existing bad actor detection sequential methods suffer from two major challenges: 
(1) Existing models rely on the RNN and its variants or transformer-based BERT~\cite{devlin2018bert} to process posts one by one, thus neglecting to recognize and leverage two different levels of information -- the post level and the sequence level (Challenge 1).
(2) When designing bad actor detectors, existing research usually did not consider the setting where bad actors are able to write adversarial posts to bypass the classifier~\cite{noorshams2020ties}, therefore making the model vulnerable in real-world applications (Challenge 2).

To bridge these gaps, we create an end-to-end Transformer-based Adversary-aware Local-Global Attended bad actor detection sequential model, called \method{}, to robustly predict labels of user sequences. 
The first component of \method{} is the local-global attended module, which leverages the transformer encoder block to encode each post bidirectionally, thus building a comprehensive post embedding. 
Next, it adopts the transformer decoder block to model the sequence of post embeddings by attention mechanism to generate the sequence embedding.
This module effectively captures the local post and the global sequence information and improves the model understanding capability. Thus, \method{} can 
make a consistent prediction of the user
even if the user post sequence has changed (Solution to Challenge 1).
The second component is the adversary-aware module, where the sequence embeddings of original sequences and modified sequences by mimicked attackers are fed into a contrastive-learning-enhanced classification layer for sequence prediction. 
The added modified sequences make \method{} knowledgeable about the adversarial changes during the training stage such that the model can recognize potential input modifications when making predictions. 
This enhanced knowledge finally makes \method{} stable against adversarial attacks that could previously fool a bad actor detector (Solution to Challenge 2).

We extensively evaluate the classification robustness of our model to show its superiority. 
Particularly, we employ two popular datasets: Yelp fake reviewer dataset~\cite{rayana2015collective} and Wikipedia vandal editor dataset~\cite{kumar2015vews}, both with ground truth bad actors. 
We compare three popular bad actor detection sequential models: TIES, a model that is used in production at Facebook~\cite{noorshams2020ties}, HRNN, a sequence classification model that uses sequential text embedding~\cite{lee2016sequential}, and one advanced transformer-based BERT model adapted for sequence classification~\cite{devlin2018bert}.
We also add two defense-involved classification models: Fine-tuning-based defense~\cite{chhabra2019data}, and Mixup Data Augmentation-based defense~\cite{zhang2017mixup}.
The experiments demonstrate that:
First, our model outperforms all compared methods in the F1 score with the highest F1 score.
Second, under the state-of-the-art next post attack by PETGEN~\cite{he2021petgen} and Large Language Model (e.g., LLaMA~\cite{touvron2023llama}), our model has the lowest decrease in F1 score.

In summary, our main contributions are:
\begin{itemize}
    \item We propose a transformer-based local-global attended module to capture both the post and the sequence information to comprehensively understand the sequence.
    \item We create a contrastive-learning-based adversary-aware training module for user classification, thus enhancing the model knowledge for potential adversarial attacks.
    \item Extensive experiments show that our method can outperform representative compared methods with the highest F1 score. When under state-of-the-art attacks, our model reports the lowest drop in F1 score. 
\end{itemize}

\section{Related Works}

\subsection{Deep Sequence Embedding-based Classification Models for Bad Actor Detection}

To identify whether a user is malicious or not, existing methods usually aim at building deep sequence embedding models to encode the sequential data and utilize the embedding for downstream applications~\cite{zhao2019recurrent,Kumar2019PredictingDE, lee2016sequential, esposito2023detecting}. 
For instance, Meta/Facebook creates a temporal embedding from users' sequence of posts, then updates users' dynamic embedding when users write a new post, and finally leverages these embeddings to identify fake accounts~\cite{noorshams2020ties}. 
On the other hand, other researchers formulate this classification of sequential posts as a sequential text classification task~\cite{lee2016sequential,  yang2016hierarchical} or a document classification problem~\cite{yang2016hierarchical}. 
Researchers first utilize Convolutional Neural Networks and RNN to capture the sequential reliance between textual posts and encode the text features for detection~\cite{lee2016sequential}. Later, researchers turn to transformer-based~\cite{vaswani2017attention} architectures due to the superior performance~\cite{dai2022revisiting, beltagy2020longformer}. 
However, when designing the classifiers, in the past researchers usually have not consider the setting where bad actors can write adversarial text to bypass the classifier, thus making the model vulnerable in real-world applications.

\subsection{Adversarial Attack and Defense on Deep Sequence Embedding-based Classification Models}

Exploration of adversarial attacks on deep sequence classifiers is a critical component to improve model robustness~\cite{zhang2020adversarial}.
Current research on attacking deep sequence embedding-based classification models primarily focuses on two types of strategies:
(1) \textit{Modification-based attacks}: These methods involve minor alterations to elements within a sequence, such as modifying the text of a post by changing or adding characters, words, or phrases. This approach, exemplified by changing the content of posts by malicious users, has been explored in various studies~\cite{ebrahimi2017hotflip, Wallace2019Triggers, li2018textbugger, jia2019towards}. 
However, these techniques often fall short because they fail to utilize the comprehensive history of a user's posts, cannot create original content, and their modifications are prone to detection through the identification of misspelled words and awkwardly manipulated sentences~\cite{pruthi2019combating}.
(2) \textit{Generation-based attacks}: This more sophisticated approach involves creating new items to append to an existing sequence, thereby crafting a new user sequence. For example, in text sequences, researchers have shown how to generate new text to fulfill attack objectives using various text generation models, such as TextGAN~\cite{nie2018relgan}, Malcom~\cite{le2020malcom}, and PETGEN~\cite{he2021petgen}. 
Some studies also employ the tactic of using post sequences from different users to generate adversarial text for attack~\cite{chiang2023shilling}.
These strategies reveal the susceptibility of deep sequence embedding-based classifiers to adversarial attacks. Nevertheless, it is noteworthy that despite these advancements in attack methodologies, there is a lack of emphasis on developing defensive mechanisms to enhance the robustness of sequential detection models against such attacks~\cite{he2021petgen}.

To defend machine learning models against adversarial attacks, the prevalent strategy involves employing min-max optimization. This approach aims to minimize the maximum adversarial loss, which represents the worst-case scenario, by computing it with adversarial examples to bolster the robustness of deep learning models~\cite{wang2021adversarial}. Such methods have been a cornerstone in enhancing model resilience across various domains, including computer vision and natural language processing (NLP).
Particularly, in the field of computer vision, adversarial examples are typically generated using techniques such as the Fast Gradient Sign Method~\cite{goodfellow2014explaining}, Projected Gradient Descent~\cite{athalye2018obfuscated}, or through the use of Generative Adversarial Networks~\cite{samangouei2018defense}. These methods have been effective in altering classification outcomes by introducing subtle, often imperceptible changes to the input data.
Conversely, in NLP, adversarial examples are crafted through various means, including the replacement of characters or words in the input text, or by adding noise to input token embeddings~\cite{liu2019multi, morris2020textattack}. These techniques aim to introduce or modify input data in a manner that can be learned by the models to improve the robustness.
Despite these advancements, directly applying these adversarial defense methods to deep sequence embedding-based classifiers poses a challenge. This is because these classifiers operate on sequential data, where attackers carefully design the subsequent entity in the sequence, rather than merely modifying existing elements. 
This distinct nature of sequential data necessitates the development of specialized defense mechanisms tailored to protect against attacks specifically designed for sequence-based models.
\section{Problem Definition}

In this section, we formally define our problem as follows:

\textbf{Preliminaries:} 
We are given $N$ users ${U} = \{u_1, ... u_N\}$ and a set of user ground truth labels $\mathcal{Y} = \{y_u\}$, 
where $y_u=0$ means user $u$ is a benign actor and $y_u=1$ means $u$ is a bad actor. 
For each user $u$, we are given a sequence of chronologically ordered posts $\matr{P}_u^{1:T}=\{\vect{p}_u^1,...,\vect{p}_u^t,..., \vect{p}_u^T \}, \matr{P}_u^{1:T} \in \mathcal{R}^{T \times d} $ where $\vect{p}_u^t \in \mathcal{R}^d $ denotes user $u$'s post at time $t$ and $d$ is the number of tokens in the post.

\textbf{Our Goal:}
We aim to build a deep user sequence embedding-based classification model $\mathcal{F}$, which can accurately generate user $u$'s predicated label $\mathcal{F}(\matr{P}_u^{1:T})$ such that $\mathcal{F}(\matr{P}_u^{1:T}) = y_u, \forall u \in {U}$ 
under the \textbf{attack setting}. 
For the attack setting, given user $u$'s sequence of posts $\matr{P}_u^{1:T}$, an attacker can use the off-the-shelf text generation model (e.g., LLAMA~\cite{touvron2023llama}, ChatGPT~\cite{wu2023brief}, PETGEN~\cite{he2021petgen}) to generate next post ${\vect{p}}_u^{T+1}$, which may flip the prediction result of the classifier on the user 
by concatenating the new post to the existing sequence (i.e., the user has a new post sequence $[{\matr{P}}_u^{1:T}, {\vect{p}}_u^{T+1}]$). 
However, the classifier is supposed to still accurately predict the label of the user as $\mathcal{F}([{\matr{P}}_u^{1:T}, {\vect{p}}_u^{T+1})]) = {y}_u$. 
We list the symbols in Table~\ref{tab:symbol}.

\begin{table}[!htbp]
    \centering
    \begin{tabular}{c|c} 
    \hline
         Notation & Description  \\
         \hline
         $\vect{p}_u^t $    & User $u$'s post at time $t$ \\
         $\matr{P}_u^{1:T}$     & User $u$'s sequence of past $T$ posts \\
         ${\vect{p}}_u^{T+1} $    & User $u$'s generated post at time $T+1$ \\
         $y_u$ & The ground truth label of user $u$  \\
         $\mathcal{F}$ & The deep user sequence embedding-based classifier \\
         $W_p$ & The position embedding matrix \\
         $W_e$ & The token embedding matrix \\
         $Emb_u$ & The sequence embedding of user $u$ \\
          \hline
    \end{tabular}
    \caption{Table of notations used in the paper. }
    \label{tab:symbol}
\end{table}

\section{Methodology}

\subsection{System Overview}

\begin{figure}[!htbp]
    \centering
    \includegraphics[width=0.8\linewidth]{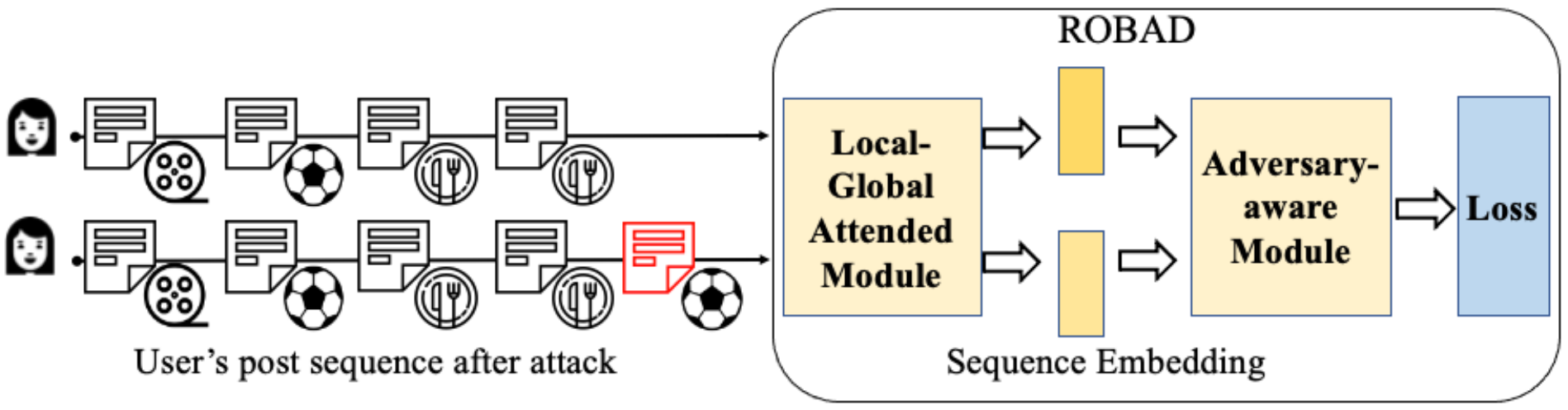}
    \caption{Overview of the \method{} architecture.}
    \label{fig:overview}
\end{figure}

In this work, we propose a Robust Adversary-aware Local-Global Attended Bad Actor Detection sequential model, called \method{}. 
Specifically, given the input of the user's historical post sequence, \method outputs the predicted label. 
\method{} has two major modules: in the first module, it leverages transformer encoder and decoder blocks to generate the local-global attended sequence embedding. 
In the second module, it utilizes contrastive learning to enhance the model capability by distinguishing the embedding of original users from the embedding of mimicked attackers.  
Through this design, \method{} can robustly predict the label of a given sequence when under attack. 
The overview of the system is shown in Figure~\ref{fig:overview}.

\subsection{Local-Global Attended Module}

\begin{figure}[!htbp]
    \centering
    \includegraphics[width=0.75\columnwidth]{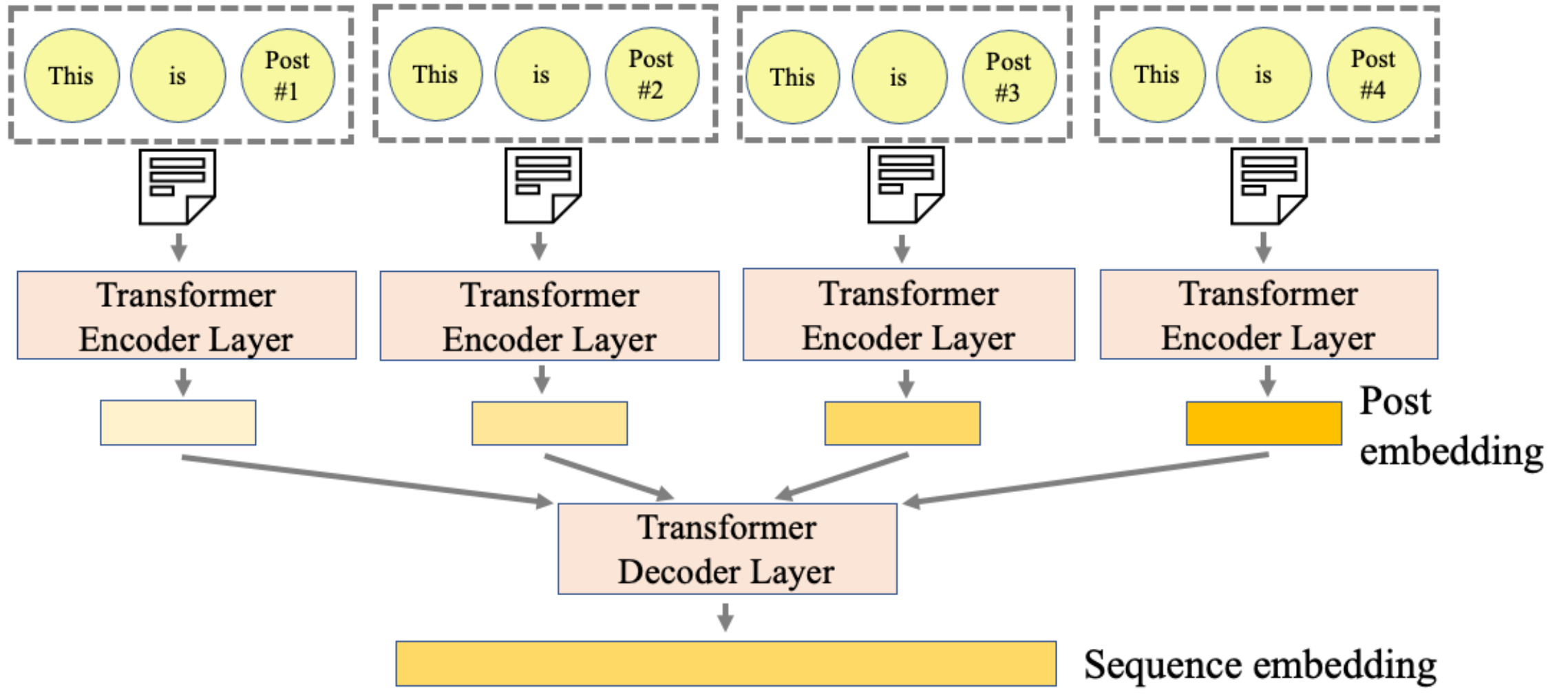}
    \caption{Local-Global attended module in \method{}.}
    \label{fig:dual_transformer}
\end{figure}

In this module, the goal is to generate the local-global attended embedding such that the embedding can comprehensively represent both the local post and global sequence information of the user. 
To achieve it, motivated by the self-attention mechanism in transformer~\cite{vaswani2017attention}, we build the encoder and decoder-based dual transformer architecture, as shown in Figure~\ref{fig:dual_transformer}. 
Finally, \method{} generates the sequence embedding of a user's sequence of historical posts.

\subsubsection{Transformer Encoder-based Component at the Local Post Level}

To represent the information in each post, we need to fully understand the post by encoding the text bi-directionally rather than unidirectionally. 
Building on the bidirectional property of transformer encoder block~\cite{vaswani2017attention}, we first pass each post $p_u^t$ to the embedding layer represented by $W_e$ which means the token embedding matrix. Then, we add the token embedding and the position embedding matrix $W_p$ to form the initial token representation $h_0^i$. 
Finally, we pass $h_0^i$ to transformer encoder blocks through $n$ layers and obtain the post embedding $h_l^i$ as the final post embedding (Note that the value of hyperparameter $n$ is examined in Section~\ref{sec:parameter_tuning}). 
This process is mathematically formulated as: 
\begin{equation}
 \begin{split}
h^i_0 &= \vect{p_u^i} W_e + W_p \\
h^i_l &= \text{transformer\_encoder\_block}(h^i_{l-1}) , \forall l \in [1, n] \\
\end{split}    
\end{equation}

\subsubsection{Transformer Decoder-based Component at the Global Sequence Level}

After passing a sequence of post $\{\vect{p}_u^1,...,\vect{p}_u^t,..., \vect{p}_u^T \}$ to the transformer encoder-based module, we obtain a list of post embeddings, $h_l^1, h_l^2, ..., h_l^i, ...$. 
%
To capture the sequential pattern in it, we need to chronologically process the posts and attend to the informative posts in the sequence. 
Motivated by the transformer decoder design where masked self-attention is deployed to model the sequential relationship and emphasize the critical entity in a sequence~\cite{vaswani2017attention}, we adopt this approach. 
Particularly, we first create a matrix $H = [h_l^1, h_l^2, ..., h_l^i]$ to represent the list of post embeddings. After adding the position embedding matrix $W_p$, we have the initial matrix $H_0$. 
Finally, we pass $H_0$ to transformer decoder blocks through $n$ layers and obtain the sequence embedding of the user $Emb_u$ (Note that the value of hyperparameter $n$ is examined in Section~\ref{sec:parameter_tuning}).
This process is formulated as:
\begin{equation}
 \begin{split}
& H_0 = H + W_p \\
& H_l = \text{transformer\_decoder\_block}(H_{l-1}), \forall l \in [1, n] \\
& Emb_u = H_l \\
\end{split}   
\end{equation}

\subsection{Adversary-aware Module}

\begin{figure}[!htbp]
    \centering
    \includegraphics[width=0.75\columnwidth]{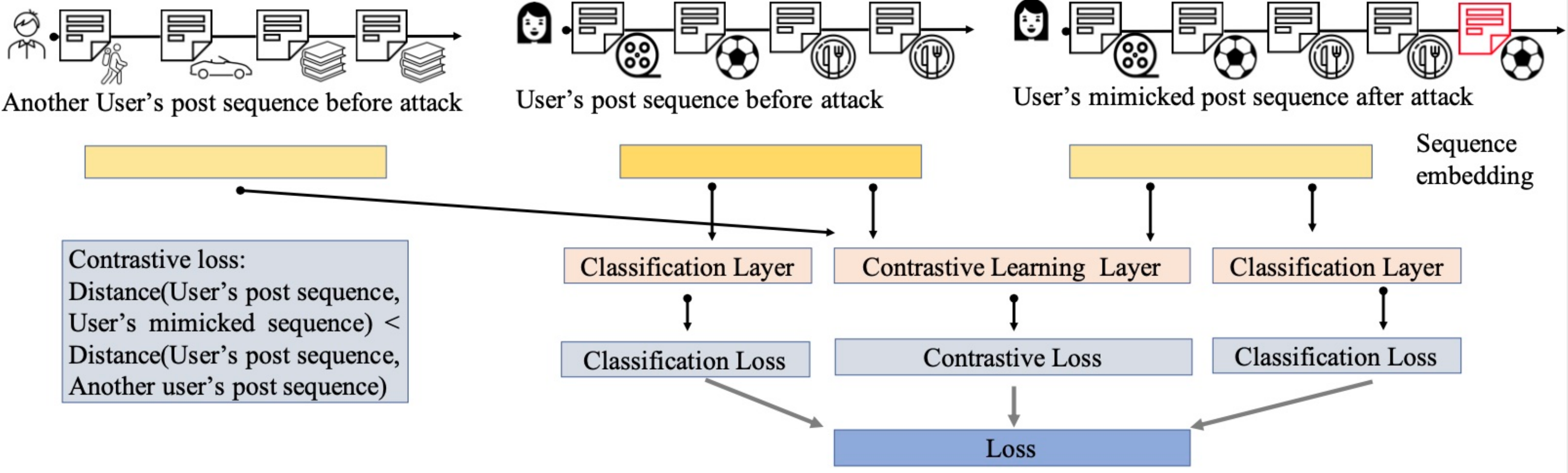}
    \caption{Adversary-aware module in \method{}.}
    \label{fig:adversarial_learning}
\end{figure}

\subsubsection{Classification Loss}

After passing a user post sequence $\{\vect{p}_u^1,...,\vect{p}_u^t,..., \vect{p}_u^T \}$ to the local-global attended module, we obtain the user embedding $Emb_u$.
Similarly, for the modified user post sequence by adversarial attacks $[P_u^{1:T}, p_u^{T+1}]$, we have $Emb_u^{attack}$.

To classify the user sequence, we pass the embedding vector through a linear layer and add a softmax function on top of the embedding as follows:
\begin{equation}
    p(y_u|Emb_u) = softmax(W Emb_u)
\end{equation}
where $W$ is the weight matrix for the linear layer. 
For the classification task, we deploy the cross entropy as the loss function:
\begin{equation}
    \mathcal{L}_{CE} =-\frac{1}{N}\sum_{i=1}^{N} y_u \log p(y_u|Emb_u) + (1 - y_u)\log (1 - p(y_u|Emb_u))
\end{equation}

We have a similar loss function for the modified post sequence by attackers as follows:

\begin{equation}
   \mathcal{L}^{attack}_{CE} =   -\frac{1}{N}\sum_{i=1}^{N} y_u \log p(y_u|Emb_u^{attack})  + (1 - y_u)\log (1 - p(y_u|Emb_u^{attack}))
\end{equation}

Finally, for the classification loss, we add the two loss functions together:

\begin{equation}
    \mathcal{L}_{CLF} = \mathcal{L}_{CE} + \mathcal{L}^{attack}_{CE}
\end{equation}

\subsubsection{Contrastive Loss}

To enhance the knowledge of models against adversarial attacks, an intuitive strategy is to leverage the mimicked behaviors of attackers during training such that the model is aware of what the attack looks like and able to recognize the attack during testing.
Following similar works in this line of adversarial machine learning~\cite{miyato2016adversarial}, we aim to ensure that the sequence-level representations of a user's original sequence and its adversarial counterpart are as similar as possible. 
In practice, this approach seeks to make the model more resistant to noise by minimizing discrepancies between genuine and adversarially modified sequences. 
Conversely, sequences that do not form a paired relationship should be distinctly separated in the representation space, enhancing the model's ability to discriminate between authentic and manipulated inputs.
To pursue this goal, we incorporate contrastive learning as a supplementary regularization technique during training. Recent advancements in contrastive learning, exemplified by MoCo~\cite{he2020momentum} and SimCLR~\cite{chen2020simple}, utilize a variety of data augmentation methods, such as random cropping and color distortion, to generate positive pairs for training. MoCo introduces a queue mechanism for managing a pool of negative examples, whereas SimCLR leverages in-batch sampling to gather negative examples. These methods train models by optimizing the InfoNCE loss~\cite{oord2018representation}, a strategy that fosters learning by distinguishing between similar (positive) and dissimilar (negative) pairs of data points. 
In \method{}, we adopt the SimCLR framework for generating positive and negative pairs, along with its loss function, to implement a contrastive learning objective that is tailored to our specific setting. 
The reason for choosing SimCLR is that SimCLR allows us to effectively model the desired relationships between original and adversarial sequences, thereby improving the noise resilience of our deep learning models.

Particularly, we first pass the sequence embedding to a non-linear projection layer for the sequence representation transfer as follows:
\begin{equation}
    \begin{split}
        z_u & = W_2 ReLU (W_1 Emb_u) \\
        z_u^{attack} & = W_2 ReLU (W_1 Emb_u^{attack}) \\
    \end{split}
\end{equation}
Where $W_1$ and $W_2$ are the weight matrix.

With a batch of $N$ user sequence examples and their corresponding adversarial examples, for each positive pair ($z_u$ and $z_u^{attack}$), there are $2(N-1)$ negative pairs, i.e., all the rest of the examples in the batch are negative examples. Here, we use $z_i$ to denote one in the $2(N-1)$ examples and have $z_u$ and $z_i$ as the negative pair. 
The contrastive learning objective is to identify the positive pair and we compute the contrastive loss using the InfoNCE loss as:
\begin{equation}
     \mathcal{L}_{InfoNCE} = -\log \left( \frac{\exp(\text{sim}(z_{u}, z_u^{attack}))}{\sum \exp(\text{sim}(z_{u}, z_{i}))} \right)
\end{equation}
where \( \text{sim}(u, v) = \frac{u^Tv}{||u||_2 ||v||_2} \) denotes the cosine similarity between two vectors.

Finally, we perform a multi-task learning diagram and take a weighted average of the classification loss and the contrastive loss as:
\begin{equation}
\mathcal{L} = w_{\text{contrastive}} \cdot \mathcal{L}_{\text{InfoNCE}} + (1 - w_{\text{contrastive}}) \cdot 
\mathcal{L}_{CLF}
\end{equation}
where \( w_{\text{contrastive}} \) indicates the weight of contrastive loss in the final loss computation.

When training the model, we minimize the loss for optimization through back-propagation using the Adam optimizer~\cite{kingma2014adam}.

\section{Experimental Evaluation}

In this section, we conduct extensive experiments to examine the performance of \method{}. Specifically, we aim to answer the following Research Questions:
\begin{itemize}
    \item \textbf{RQ1}: Is \method{} able to effectively identify bad actors on online platforms? 
    \item \textbf{RQ2}: When under attack, can \method{} successfully defend against adversarial attacks and output the correct predictions?
    \item \textbf{RQ3}: What is the contribution of each module in \method{} towards its performance?
    \item \textbf{RQ4}: How sensitive is the performance of \method{} when we change its major parameters?
\end{itemize}

\subsection{Datasets}
\begin{table}[!htbp]

\begin{tabular}{lrr}
\toprule
               Dataset  &  Wikipedia & Yelp\\
\midrule
          Number of users  &        794 &  3,940\\
   Number of benign users  &        397 &  2,016\\
Number of malicious users  &        397 &  1,924\\
    Total number of posts  &      11,547& 35,123 \\
    Median posts per user  &         15 &     9\\
\bottomrule
\end{tabular}
\caption{Dataset Statistics}
\label{tab:data-statistics}
\end{table}

We conducted our evaluation using real-world datasets from two widely-used platforms: Wikipedia and Yelp, with their detailed statistics presented in Table~\ref{tab:data-statistics}. \\
(a) \textbf{Wikipedia dataset:} Comprising user (or editor) contributions to Wikipedia articles, this dataset differentiates between benign and vandal editors, the latter of whom were identified and subsequently removed by Wikipedia administrators~\cite{kumar2015vews}. Each user's contribution history is recorded as a sequence of edits (i.e., an edit is a post in the context) made to articles, providing a basis for analysis. \\
(b) \textbf{Yelp dataset:} This dataset captures Yelp users' reviews of restaurants, distinguishing between benign reviewers and fraudulent ones. Fraudulent reviewers are flagged by Yelp's internal classification algorithms~\cite{rayana2015collective}. Here, a user sequence is composed of reviews (i.e., a review is a post in the context) made by the user, offering insights into their review patterns.

For both datasets, we ensured enough information in each user's sequence by: users with fewer than 5 posts and posts containing fewer than 5 tokens were excluded. 
To construct a user sequence, we utilized the most recent 20 posts from each user, aiming to represent the most current user behavior.

\subsection{Evaluation Metrics}\label{sec:evaluation_metric}
To evaluate \method{}, we deploy several metrics to measure the following:\\
\noindent (a) \textbf{Classification Performance:} Following the conventional measurement of classification models, we report the precision, recall, and F1-score.

\noindent (b) \textbf{Robustness Performance:} Similar to other adversarial attack and defense works~\cite{he2021petgen}, we report the F1 score after the attack, and compare it with the F1 score before the attack. If the F1 score drops considerably after the attack, the attack is then successful.

\subsection{Compared Models}
In this work, we compare \method{} with three popular deep user sequence classification models and two defense models against adversarial attacks. \\
(1) \textbf{Hierarchical Recurrent Neural Network (HRNN)}~\cite{zhao2019recurrent} captures the sequential pattern of the input text by the hierarchical neural network structure for accurate classification. In HRNN, each user post is first transformed into a vector, and the sequence of user post vectors is converted into a compact user embedding, which is finally used for user classification. \\
(2) \textbf{Temporal Interaction EmbeddingS (TIES)}~\cite{noorshams2020ties} is developed and deployed by Facebook/Meta to detect malicious accounts. To adapt it in our sequence classification setting, we utilize the temporal embedding component of the TIES model for classification (as there is no graph structure in our datasets). Specifically, TIES converts a user post sequence into a user embedding vector by a sequence encoding layer and a pooling layer. 
(3) \textbf{Feature-based BERT (F-BERT)}~\cite{devlin2018bert} is a feature-based approach utilizing the advances of bidirectional transformers to encode and represent text. Here, to adapt to the sequence classification setting, we first obtain the BERT embedding for each post and pass the sequence of embeddings to another transformer layer for classification. \\ 
(4) \textbf{Fine-Tuning-based defense (FT-Defense)}~\cite{chhabra2019data}: In this approach, we deploy the multi-stage fine-tuning approach based on adversarial attack data to improve the model robustness. Specifically, we first use the original data to train the model and have a frozen reference model. Next, we feed the adversarial attack data to fine-tune the model. In this step, for each data point, we have a prediction difference loss from the frozen reference model and active fine-tuned model to refine the model through back-propagation. Then, we repeat the fine-tuning process until the model converges. \\ 
(5) \textbf{Mixup Data Augmentation-based defense (MDA-Defense)}~\cite{zhang2017mixup}: Instead of using the original and attack data in different steps as shown in the FT-defense, this method will combine the two sets of data by using the widely deployed mixup data augmentation strategy and employ the combined data points to train the model. 
\\

\subsection{Experiment Setup}
During the experiment, we split the dataset by five-fold cross-validation and report the average numbers. 
The number of tokens in a post is $30$, and the learning rate is $1e-3$. 
We use Adam as the optimizer with the mini-batch size of $32$~\cite{Jeff2017Ian}. 
For the transformer encoder and decoder blocks, we set the embedding size and the number of attention heads as $128$ and $2$.

\subsection{RQ1: Effectiveness of Classifiers}
In this section, we evaluate \method{} on Wikipedia and Yelp datasets. The classification result is presented in Table~\ref{tab:rq1-wikipedia} and Table~\ref{tab:rq1-yelp}.
%
\begin{table}[!htbp]
\begin{tabular}{|c|c|c|c|c|}
\hline
\multicolumn{2}{|c|}{Method} & Precision & Recall & F1 \\ \hline
\multirow{3}{*}{Sequence-based} & HRNN & 0.652 & 0.667 & 0.658 \\ \cline{2-5}
& TIES & \cellcolor{blue!25} 0.689 & 0.645 & 0.666 \\ \cline{2-5}
& F-BERT & 0.553 & 0.619 & 0.582 \\ \hline
\multirow{2}{*}{Defense-involved} & FT-Defense & 0.627 & \cellcolor{blue!10}0.746 & \cellcolor{blue!10}0.682 \\ \cline{2-5}
& MDA-Defense & \cellcolor{blue!10}0.683 & 0.667 & 0.671 \\ \hline
\multicolumn{2}{|c|}{\method{} (Ours)}& 0.615 & \cellcolor{blue!25}0.820 & \cellcolor{blue!25}0.701 \\ \hline
\end{tabular}
\caption{Comparison of classification performance by different models on Wikipedia dataset (RQ1).}
\label{tab:rq1-wikipedia}
\end{table}

\begin{table}[!htbp]
\begin{tabular}{|c|c|c|c|c|}
\hline
\multicolumn{2}{|c|}{Method} & Precision & Recall & F1 \\ \hline
\multirow{3}{*}{Sequence-based} & HRNN & 0.659 & 0.684 & 0.675 \\ \cline{2-5}
& TIES & \cellcolor{blue!10} 0.680 & 0.687 & 0.683 \\ \cline{2-5}
& F-BERT & 0.565 & 0.646 & 0.601 \\ \hline
\multirow{2}{*}{Defense-involved} & FT-Defense & 0.642 & 0.661 & 0.656 \\ \cline{2-5}
& MDA-Defense & 0.677 & \cellcolor{blue!10} 0.700 & \cellcolor{blue!10} 0.688 \\ \hline
\multicolumn{2}{|c|}{\method{} (Ours)}& \cellcolor{blue!25}0.707 & \cellcolor{blue!25} 0.710 & \cellcolor{blue!25}  0.708 \\ \hline
\end{tabular}
\caption{Comparison of classification performance by different models on Yelp dataset (RQ1).}
\label{tab:rq1-yelp}
\end{table}

As we can see, \method{} has the highest F1 score among all compared methods on two datasets, showing its superiority. The reason may be that \method{} deploys the advanced transformer encoder and decoder architectures and increases the training data through adversarial machine learning. 
Particularly, compared to sequence-based solutions, \method{} has the highest precision, recall, and F1 score in most cases except the precision is slightly lower on the Wikipedia dataset. The potential reason is the smaller data size of the Wikipedia dataset, which makes the training of \method{} less satisfactory. When trained and tested on the larger Yelp dataset, \method{} beats all sequence-based approaches on all metrics of precision, recall, and F1 score.
Second, when comparing \method{} with the defense-involved approaches, we have similar findings -  \method{} is the best in most cases, especially on the larger Yelp dataset.

\subsection{RQ2: Robustness of Classifiers}

To evaluate the robustness of classifiers under adversarial attacks, we first mimic the behavior of malicious users by creating a new post using text generation models. Next, we concatenate the new post to the corresponding post sequence and feed it into the classification again. We report the  F1 score after the attack as mentioned in Section~\ref{sec:evaluation_metric}.
In practice, we leverage the state-of-the-art text generation attack method against the deep sequence embedding-based classifiers (i.e., PETGEN~\cite{he2021petgen}) to mimic the attack behaviors. 
On the other hand, due to the advance of large language models (LLM), we also deploy LLM to generate the adversarial text for attack. In our experiment, we select the widely-used LLaMA~\cite{touvron2023llama} model.
The F1 score after attack on Wikipedia and Yelp datasets are reported in Table~\ref{tab:robustness-wikipedia} and Table~\ref{tab:robustness-yelp}, respectively.
%


\begin{table}[!h]
\begin{tabular}{|c|c|c|c|c|}
\hline
\multicolumn{2}{|c|}{Method} & Without Attack & After PETGEN Attack & After LLaMA Attack \\ \hline
\multirow{3}{*}{Sequence-based} & HRNN & 0.658 &   0.605 & 0.628 \\ \cline{2-5}
& TIES & 0.666 & 0.591 & 0.581 \\ \cline{2-5}
& F-BERT & 0.582 & 0.573 & 0.570 \\ \hline
\multirow{2}{*}{Defense-involved} & FT-Defense &  \cellcolor{blue!10} 0.682 & 0.620 & 0.631 \\ \cline{2-5}
& MDA-Defense & 0.671 & \cellcolor{blue!10}0.641 &  \cellcolor{blue!10} 0.665 \\ \hline
\multicolumn{2}{|c|}{\method{} (Ours)}& \cellcolor{blue!25} 0.701 & \cellcolor{blue!25} 0.695 & \cellcolor{blue!25} 0.697 \\ \hline
\end{tabular}
\caption{Comparison of robustness performance by different models on Wikipedia dataset (RQ2). Here, we report the F1 score after the attack.}
\label{tab:robustness-wikipedia}
\end{table}

\begin{table}[!h]
\begin{tabular}{|c|c|c|c|c|}
\hline
\multicolumn{2}{|c|}{Method} & Without Attack & After PETGEN Attack & After LLaMA Attack \\ \hline
\multirow{3}{*}{Sequence-based} & HRNN & 0.675 & 0.641 & 0.628 \\ \cline{2-5}
& TIES & 0.683 & 0.661 & 0.659 \\ \cline{2-5}
& F-BERT & 0.601 & 0.587 & 0.571 \\ \hline
\multirow{2}{*}{Defense-involved} & FT-Defense & 0.656 & 0.650 & 0.652 \\ \cline{2-5}
& MDA-Defense &  \cellcolor{blue!10} 0.688 &  \cellcolor{blue!10} 0.673 &  \cellcolor{blue!10} 0.664 \\ \hline
\multicolumn{2}{|c|}{\method{} (Ours)}& \cellcolor{blue!25} 0.708 & \cellcolor{blue!25} 0.700 & \cellcolor{blue!25} 0.682 \\ \hline
\end{tabular}
\caption{Comparison of robustness performance by different models on Yelp dataset (RQ2). Here, we report the F1 score after the attack.}
\label{tab:robustness-yelp}
\end{table}

\begin{table}[!h]
\begin{tabular}{|c|c|c|c|c|}
\hline
\multicolumn{2}{|c|}{Method} & Without Attack & After PETGEN Attack & Relative drop in F1 score \\ \hline
\multirow{3}{*}{Sequence-based} & HRNN & 0.675 & 0.641 & 5.037\% \\ \cline{2-5}
& TIES & 0.683 & 0.661 & 3.221\% \\ \cline{2-5}
& F-BERT & 0.601 & 0.587 & 2.329\% \\ \hline
\multirow{2}{*}{Defense-involved} & FT-Defense & 0.665 & 0.650 & 2.255\% \\ \cline{2-5}
& MDA-Defense &   0.688 &   0.673 &   2.181\% \\ \hline
\multicolumn{2}{|c|}{Our model}& \cellcolor{blue!25} 0.708 & \cellcolor{blue!25} 0.700 & \cellcolor{blue!25} 1.129\%\\ \hline
\end{tabular}
\caption{Comparison of robustness performance by different models on Yelp dataset (RQ2). Here, we report the F1 score after the attack.}
\label{tab:robustness-yelp}
\end{table}

The results show that across various attack settings, \method{} has the lowest decrease in F1 score and maintained the highest F1 score among all compared methods. This demonstrates the highest robustness of our \method{} method. 
Particularly, compared to the sequence-based methods, \method{} only decreases $0.713\%$ on average in F1 score while others worsen with even $12.76\%$ decrease in F1 score.
Second, when comparing with defense-involved methods, we find that \method{} still works better even if MDA-Defense is the second best with only $2.68\%$ decrease on average in F1 score. 
Interestingly, we also notice that defense-involved solutions are better than sequence-based ones when under attack. This is reasonable since defense-involved solutions explicitly take adversarial examples in the training stage to attend to the attack scenario.

\subsection{RQ3: Ablation Studies}

To examine the effectiveness of each module in \method{}, we conduct the ablation study where we test the performance of different variants of \method{}. Particularly, we examine: 
\begin{itemize}
    \item \method{}: The full model with two modules.
    \item \method{} - w/o Local-Global Attended Module: We remove the local-global attended module in \method{}.
    \item \method{} - w/o Adversary-aware Module: We remove the adversary-aware module in \method{}
\end{itemize}
The results on Wikipedia and Yelp datasets are presented in Table~\ref{tab:ablation-wiki} and Table~\ref{tab:ablation-yelp}, respectively.

\begin{table}[!htbp]
\centering
\begin{tabular}{|l|c|c|c|}
\hline
{Method} & {Without Attack} & {After PETGEN Attack} & {After LLaMA Attack} \\ \hline
\method{} & \cellcolor{blue!25} 0.701 & \cellcolor{blue!25} 0.695 & \cellcolor{blue!25} 0.697 \\ \hline
- w/o Local-Global Attended Module & 0.673 & 0.661 & 0.623 \\ \hline
- w/o Adversary-aware Module & 0.690 & 0.652 & 0.619 \\ \hline
\end{tabular}
\caption{Ablation studies of \method{} on Wikipedia dataset (RQ3), measured by F1 score.}
\label{tab:ablation-wiki}
\end{table}
\begin{table}[!htbp]
    \centering
    \begin{tabular}{|l|c|c|c|}
    \hline
    {Method} & {Without Attack} & {After PETGEN Attack} & {After LLaMA Attack} \\ \hline
    \method{} & \cellcolor{blue!25} 0.708 & \cellcolor{blue!25} 0.700 & \cellcolor{blue!25} 0.682 \\ \hline
    - w/o Local-Global Attended Module & 0.664 & 0.652 & 0.642 \\ \hline
    - w/o Adversary-aware Module & 0.678 & 0.647 & 0.631 \\ \hline
    \end{tabular}
    \caption{Ablation studies of \method{} on Yelp dataset (RQ3), measured by F1 score.}
    \label{tab:ablation-yelp}
\end{table}

As shown in the results, \method{} containing two modules always performs the best among all other variants when measured by the F1 score. When removing any module, the performance drops. This demonstrates the contribution of each module and the necessity of having two modules work together.
Particularly, when comparing different variants, we find that removing the local-global attended module leads to $4.443\%$ decrease on average while removing the adversary-aware module causes $3.878\%$ decrease. 
This is potentially because the local-global attended module increases the model predictive capability while the adversary-aware module improves the model knowledge against adversarial attacks. 
In our scenario of the user sequence classification, the local-global attended module weighs more since the predictive capability is the building block of a well-performed system.

\subsection{RQ4: Hyperparameter Tuning}\label{sec:parameter_tuning}
In this section, we examine the effect of different hyperparameters on the model performance. Here, we focus on three major parameters: the number of layers in the transformer encoder, the number of layers in the transformer decoder, and the weight of the contrastive loss $w_{contrastive}$. We report the F1 score when the model is without attack (Note that, we have similar findings when the model is under attack. However, we omit them in this section to save space.)

\begin{table}[!htbp]
    \centering
    \begin{tabular}{|c|c|c|c|}
    \hline
    {\# of layers} & 1 & 2 & 3 \\ \hline
    Wikipedia & \cellcolor{blue!25}0.701 & 0.6859 & 0.6638 \\ \hline
    Yelp & \cellcolor{blue!25} 0.708 & 0.6586 & 0.6435 \\ \hline
    \end{tabular}
    \caption{Effects of varying number of transformer encoder layers in \method{} (RQ4), measured by F1 score.}
    \label{tab:number_encoder}
\end{table}

\begin{table}[!htbp]
    \centering
    \begin{tabular}{|c|c|c|c|}
    \hline
    {\# of layers} & 1 & 2 & 3 \\ \hline
    Wikipedia & \cellcolor{blue!25} 0.701 & 0.6790 & 0.6631 \\ \hline
    Yelp & \cellcolor{blue!25} 0.708 & 0.6569 & 0.6359 \\ \hline
    \end{tabular}
    \caption{Effects of varying number of transformer decoder layers in \method{} (RQ4), measured by F1 score.}
    \label{tab:number_decoder}
\end{table}

\begin{table}[!htbp]
    \centering
    \begin{tabular}{|c|c|c|c|c|c|c|}
    \hline
    {Weight Value} & 0 & 0.1 & 0.3 & 0.5 & 0.7 & 0.9 \\ \hline
    Wikipedia & 0.690 &  \cellcolor{blue!25} 0.701 & 0.670 & 0.636 & 0.596 & 0.584 \\ \hline
    Yelp & 0.678 & \cellcolor{blue!25} 0.708 & 0.661 & 0.622 & 0.605 & 0.591 \\ \hline
    \end{tabular}
    \caption{Effects of varying weight values of contrastive loss in \method{} (RQ4), measured by F1 score.}
    \label{tab:contrastive_weight}
\end{table}

As we can see in Table~\ref{tab:number_encoder}, for the number of transformer encoder layers, the best value is $1$ and the higher value leads to worse performance. The reason can be that more layers lead to higher model complexity. This finally causes the overfitting problem, especially given the relatively small size of the two datasets. 
A similar finding also holds for the number of transformer decoder layers as shown in Table~\ref{tab:number_decoder}.
When examining Table~\ref{tab:contrastive_weight}, we find that $0.1$ is the best value for $w_{constrastive}$. It implies that when we add the contrastive loss to the classification loss, it helps improve the classification performance.
But, the weight should not be large. $0.1$ works best in the current
setup. If we decrease the value to $0$, the model performs worse.
All these results indicate the contribution of each module. When using them together, we should take caution and exhaustively test different values to find the proper hyperparameter set for the specific application setting.
\section{Conclusion}
In this study, we introduce a novel deep learning framework for bad actor detection, specifically designed to identify malicious users and bolster defenses against adversarial attacks. Our model applies a transformer encoder block to bidirectionally encode each post, creating comprehensive post embeddings that capture nuanced textual features. Subsequently, these embeddings are processed through a transformer decoder block. This block leverages an attention mechanism to discern and model the sequential patterns inherent in the post embeddings, generating the sequence embedding for a user. 
Finally, the post embedding is passed through the contrastive learning-enhanced classification layer for the user prediction task. 
As shown in the experimental results, our solution has the lowest drop in F1 score across various attack settings when compared with various sequence-based and defense-involved competitors.

Despite its successes, our work has some limitations. 
A primary constraint is the model's current design, which only accommodates English language posts, potentially overlooking the rich diversity of global social media discourse that encompasses a multitude of languages. 
Additionally, the model's architecture is tailored exclusively for sequential data, thereby omitting the potential insights that could be gleaned from more complex data structures like graphs. This focus on sequences also means the model's defensive capabilities are specifically tuned to counter the creation of new posts by attackers, leaving room for future enhancements to address a broader spectrum of adversarial strategies, including but not limited to, more sophisticated manipulation techniques that might exploit other aspects of user behavior or platform interaction. 
Besides, we examine our method on relatively small datasets (Note that it is still the largest public dataset in the existing literature on bad actor detection sequential models). We plan to extend the data size in the future. 
These areas present fertile ground for future research, aiming to extend the model's applicability and defensive reach across a broader array of languages, data structures, and attack methodologies.


\bibliographystyle{ACM-Reference-Format}
\bibliography{main}

\appendix

\end{document}